\documentclass[runningheads]{llncs}
\usepackage{xcolor}          
\usepackage[colorlinks=true,   
            citecolor=blue,    
            linkcolor=blue,    
            urlcolor=blue,     
            filecolor=magenta  
]{hyperref}
\usepackage{amssymb} 
\usepackage{bbding} 
\usepackage{booktabs}    
\usepackage{multirow}    
\usepackage[T1]{fontenc}
\usepackage{graphicx,verbatim}
\begin{document}
\title{
  Parameter-Efficient Adaptation of SAM3 for Prompt-Driven Surgical Concept Segmentation
}
\titlerunning{Parameter-Efficient Adaptation of SAM3 for Surgical Segmentation} 
%

\author{Changjing Liu\inst{1,} \and
Yiming Huang\inst{1} \and
Beilei Cui\inst{1} \and
Liangjing Shao\inst{1} \and
Long Bai\inst{1} \and
Yanheng Li\inst{2} \and
Haoxuan Che\inst{2} \and
Hongliang Ren\inst{1}\thanks{Corresponding author.}}

\institute{Department of Electronic Engineering, The Chinese University of Hong Kong, Hong Kong SAR, China\\ 
\and XGEN Labs \email{\{changjingliu,yhuangdl,beileicui,leonking-shaw\}@link.cuhk.edu.hk, b.long@ieee.org, yanhengli3-c@my.cityu.edu.hk, chehaoxuan1220@gmail.com, hlren@ee.cuhk.edu.hk}}

\authorrunning{C. Liu et al.}
\maketitle              
\begin{abstract}
Efficient surgical segmentation empowers clinical diagnosis, intraoperative monitoring, and downstream robotic pipelines for reconstruction and simulation.
Although prompt-driven foundation models like Segment Anything Model 3 (SAM3) achieve strong segmentation performance on natural images, surgical data exhibits domain gaps against its pre-training data, resulting in degraded segmentation accuracy.
Furthermore, existing medical SAM methods require full-parameter fine-tuning, incurring heavy computational consumption and low efficiency.
To address these limitations, this work proposes a parameter-efficient Low-Rank Adaptation (LoRA) adaptation of SAM3 for surgical concept segmentation. 
We inject low-rank adapters into the prompt encoder, detector and tracker while fully freezing the vision backbone, which only optimizes \textbf{0.98\%} of the total model parameters and supports training on a single consumer GPU.
Comprehensive experiments demonstrate that our method consistently outperforms zero-shot SAM3 and other mainstream baselines, and the generated segmentation results can be directly deployed to support downstream  robotic surgical scene reconstruction and physical simulation pipelines. 
Our code is available at \url{https://github.com/ChangjingLiu/SurgSAM3}.

\keywords{Surgical Concept Segmentation \and Parameter-Efficient Fine-Tuning \and Low-Rank Adaptation \and SAM3.}

\end{abstract}

\section{Introduction}

\begin{figure*}[t]
  \centering
  \includegraphics[width=1\textwidth]{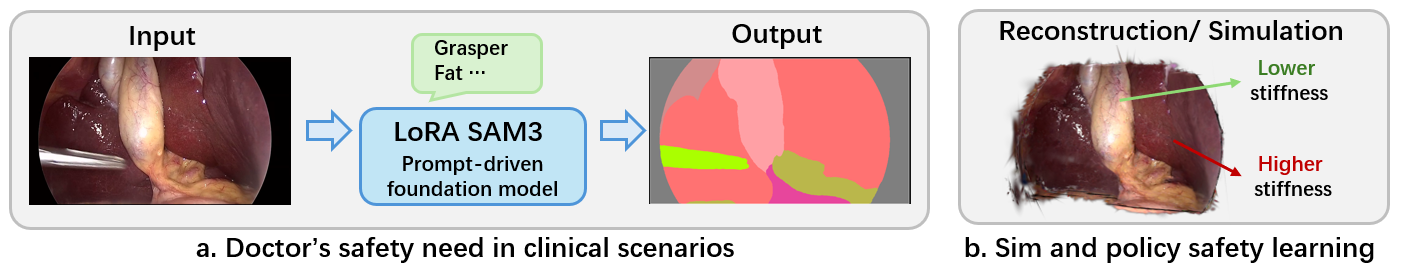}
  \caption{(a) Our efficient adaptation of SAM3 generates class-wise segmentation masks from surgical frames and prompts; (b) generated segmentation results power a downstream reconstruction and simulation pipeline for policy learning.}
  \label{fig:overview_sam3}
\end{figure*}

Efficient surgical concept segmentation is fundamental to modern clinical workflows, providing quantitative metrics for robust diagnosis, while acting as a prerequisite for downstream robotic surgery applications: scene reconstruction~\cite{kerbl20233d,wu20244d}, physics-based simulation~\cite{liu2026endogsim}, and policy learning.
Despite remarkable progress in Convolutional neural networks (CNNs)~\cite{ronneberger2015u,milletari2016v,huang2020unet} and vision transformers (ViTs)~\cite{cao2022swin,hatamizadeh2022unetr,wasserthal2023totalsegmentator} in medical segmentation, most existing models are optimized for specific data distributions, lacking generalizability to new scenes. 
This constraint highlights the demand for robust, generalizable surgical segmentation models~\cite{guan2021domain}.

Large foundation models, such as the Segment Anything Model (SAM)~\cite{ravi2025sam,kirillov2023segment} have brought a paradigm shift to universal visual segmentation.
Built on prompt-based visual modeling, SAM~\cite{kirillov2023segment} delivers strong zero-shot segmentation generalization across natural images. 
Later variants like SAM3~\cite{carion2025sam} promote this paradigm by introducing promptable concept segmentation (PCS) with open-vocabulary text and exemplar prompts.
Within medical domain, methods such as MedSAM~\cite{ma2024segment}, MedSAM2~\cite{ma2025medsam2}, Medical SAM2~\cite{zhu2024medical}, and Medical SAM3~\cite{jiang2026medical} have successfully validated this strategy by adapting vanilla SAM to medical tasks. 
However, existing medical SAM approaches are mainly trained on general medical datasets and lack dedicated optimization for surgical scenes.
In addition, these methods require full-parameter fine-tuning that consumes up to 80 GB GPU memory, making them infeasible for deployment on affordable consumer-grade GPUs in routine clinical practice.
Low-Rank Adaptation (LoRA), a representative Parameter-Efficient Fine-Tuning (PEFT) technique~\cite{hu2022lora}, resolves this constraint. 
It injects lightweight trainable rank-factorized matrices into frozen Transformer, achieving favorable cross-domain transfer using minimal data and GPU resource consumption~\cite{ding2023parameter}.

To achieve efficient cross-domain transfer for surgical imaging, we turn to lightweight parameter adaptation rather than full fine-tuning.
Specifically, we fine-tune only the prompt encoder, detector, and tracker while freezing the vision backbone, updating merely 0.98\% of model parameters. 
This efficient design enables single-GPU training and preserves the visual representation capability of SAM3 for medical applications.
Evaluated on CholecSeg8k~\cite{hong2020cholecseg8k}, EndoVis18~\cite{allan20202018}, and CaDISv2~\cite{grammatikopoulou2021cadis} with a single universal LoRA weight, our method outperforms dataset-specific SurgTPGS~\cite{huang2025surgtpgs} and fully fine-tuned Medical SAM3~\cite{jiang2026medical} under identical training settings.
Our code and weights will be released upon acceptance.
Specifically, our work comprises three core contributions:

\begin{itemize}
\item A parameter-efficient adaptation of SAM3 for surgical concept segmentation, updating only \textbf{0.98\%} of parameters while training GPU memory peaks at \textbf{9 GB}, enabling deployment on resource-limited edge hardware.
\item  A unified input representation is adopted to support shared weights across diverse surgical benchmarks without dataset-specific retraining, greatly reducing the computational cost of per-dataset fine-tuning.
\item Extensive experiments validate our method surpasses zero-shot SAM3 and other mainstream approaches, producing high-quality segmentation for downstream physics-driven reconstruction and simulation.
\end{itemize}

\section{Methods}

\begin{figure*}[t]
  \begin{center}
    \centerline{\includegraphics[width=1\columnwidth]{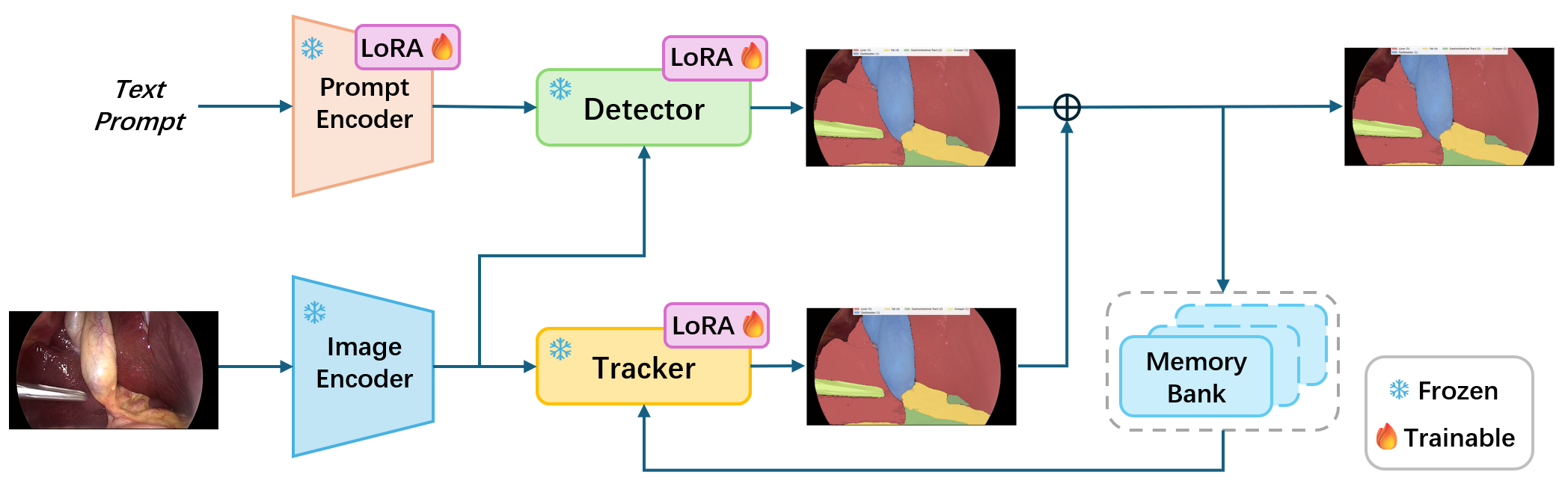}}
    \caption{
      Efficient adaptation of SAM3 for surgical segmentation via LoRA. 
      The backbone remains frozen; only \textbf{0.98\%} of parameters (low-rank adapters on the prompt encoder, detector, and tracker) are trainable.
    }
    \label{fig:workflow}
  \end{center}
\end{figure*}

\subsection{Unified Input Representation}
\label{sec:preliminaries}
To facilitate prompt-guided segmentation generalizable across heterogeneous surgical datasets and reduce training overhead, we standardize the native class labels of all datasets into a unified concept vocabulary $\mathcal{V}$.
Specifically, each surgical training sample in our pipeline is a triplet $(\mathbf{I}, \mathbf{m}_{gt}, c)$, where $\mathbf{I} \in \mathbb{R}^{H \times W \times 3}$ is an RGB frame from surgical video, $\mathbf{m}_{gt} \in \{0, 1\}^{H \times W}$ is a per-class binary ground-truth segmentation mask, and $c \in \mathcal{V}$ is a text prompt such as \emph{``grasper''}.
To keep the prompt vocabulary compatible across \textit{CholecSeg8k}~\cite{hong2020cholecseg8k}, \textit{EndoVis18}~\cite{allan20202018}, and \textit{CaDISv2}~\cite{grammatikopoulou2021cadis}, we map each dataset's native class labels to the shared concept dictionary $\mathcal{V}$ (e.g.\ \emph{Liver} / \emph{L-hook Electrocautery}).
This design supports cross-dataset prompt reuse, eliminating redundant prompt tuning and improving experimental efficiency.

\subsection{Foundation Models for Surgical Segmentation}
\label{sec:sam-arch}
SAM3~\cite{carion2025sam} is a prompt-driven foundation model well-suited for surgical segmentation. 
Built on the promptable segmentation design of SAM2~\cite{ravi2025sam}, SAM3 delivers Promptable Concept Segmentation (PCS) for surgical scenes. 
It takes a frame $\mathbf{I}$ and a concept prompt $c$ as input and outputs a set of predicted masklets $\mathcal{M} = \{(\mathbf{m}_k, s_k, p_k)\}_{k=1}^{K}$, where $\mathbf{m}_k \in \{0, 1\}^{H \times W}$ is a binary predicted mask, $s_k \in [0, 1]$ is a confidence score, and $p_k \in \mathbb{N}$ is a trackable identity.
SAM3 comprises three coupled components (Fig.~\ref{fig:workflow}): 
\textit{(i)}~an Image-prompt encoder that produces joint image and text tokens; 
\textit{(ii)}~a DETR-style detector~\cite{carion2020end} that decodes concept-conditioned box and mask proposals; 
\textit{(iii)}~a memory-bank tracker that propagates masks across frames via memory buffer $\mathcal{B}_t$.
The final prediction at frame $t$ fuses the detector proposal and the tracker-propagated masklet by confidence-weighted Non-Maximum Suppression (NMS).

\subsection{Efficient LoRA Adaptation of SAM3}
\label{sec:lora}
We adapt SAM3 to surgical scenes via Low-Rank Adaptation (LoRA)~\cite{hu2022lora}, which injects trainable low-rank updates into frozen linear projections (Fig.~\ref{fig:workflow}). 
For each frozen weight $W_0 \in \mathbb{R}^{d\times k}$, LoRA computes updated weights via low-rank decomposition: $W = W_0 + B A$ with $B\in\mathbb{R}^{d\times r}$, $A\in\mathbb{R}^{r\times k}$ and rank $r \ll \min(d,k)$.
Only $A$ and $B$ are optimized, which keeps the number of trainable parameters small and preserves the foundation model's pre-trained representations.

Following the SAM3 architecture introduced in Sec.~\ref{sec:sam-arch}, we inject LoRA only into the prompt encoder, DETR decoder and tracker, while fully freezing the vision encoder and all other components:
\textit{(i) Prompt encoder}: LoRA is applied to cross-modal alignment layers to align surgical terminology (e.g., L-hook electrocautery, kidney parenchyma) with pre-trained text embeddings $\mathbf{Z}_c$.
\textit{(ii) DETR-style detector}: We equip all attention projections $(q, k, v, out\_proj)$ with LoRA to handle rare surgical instruments unseen during pre-training.
\textit{(iii) Tracker}: Low-rank branches adapt tracking attention layers for stable cross-frame mask propagation.
We exclude the vision encoder, as its low-level visual features generalize well to surgical data without fine-tuning.

Our trainable parameters total 8.32 M, accounting for merely 0.98\% of the full 849 M SAM3 backbone parameters. 
Compared to full-parameter fine-tuning, which typically demands about 80 GB of GPU memory, our approach significantly reduces training memory consumption to only 9 GB, achieving an over 80\% reduction in memory overhead and making it well-suited for resource-constrained deployment.

\section{Experiments}

\subsection{Datasets and Implementation Details}
\label{sec:datasets}

\noindent\textbf{Datasets.}
We evaluate on three public surgical datasets containing different procedures, instruments, and tissue types: \textit{CholecSeg8k}~\cite{hong2020cholecseg8k}, \textit{EndoVis18}~\cite{allan20202018}, and \textit{CaDISv2}~\cite{grammatikopoulou2021cadis}, covering cholecystectomy, nephrectomy and cataract surgery with distinct tissue and instrument labels.
We reserve the listed videos (01\_00080, 01\_00240, 01\_15019, 12\_15750 from \textit{CholecSeg8k}; Seq\_5, Seq\_9 from \textit{EndoVis18}; videos 2, 12, 22 from \textit{CaDISv2}) as the held-out test set for evaluation, while all remaining samples are used for LoRA fine-tuning.

\noindent\textbf{Implementation Details.}
SAM3 was initialized from its publicly available pretrained weights. We use a lightweight LoRA setup ($r=16$, $\alpha=32$, dropout rate $0.1$) inserted into the prompt encoder, detector and tracker, while freezing vision encoders. 
Training adopts bf16 mixed precision with batch size $1$ and gradient accumulation $8$. 
We optimize via AdamW ($\beta_1=0.9$, $\beta_2=0.999$, $\epsilon=10^{-8}$) with weight decay $0.01$, an initial cosine learning rate of $10^{-4}$ and $200$ warmup steps, gradient clipping norm $1.0$, for $10$ epochs, selecting the checkpoint with the highest validation mIoU. 
All experiments run on one RTX 3090 consumer GPU, with peak training memory usage of only 9 GB.

\noindent\textbf{Evaluation Metrics.}
We evaluate performance on the test set via two standard medical segmentation metrics: mean Intersection-over-Union (mIoU) and Dice coefficient. 
We compute scores for each class and average them to obtain overall results for fair cross-method comparison.

\begin{table}[!t]
    \centering
    \caption{
    \textbf{Quantative results on the CholecSeg8K~\cite{hong2020cholecseg8k} dataset}.    
  We compare the mIoU scores (\%), and highlighted the \colorbox{red!40}{first}, \colorbox{orange!50}{second}, and \colorbox{yellow!50}{third}. Our method outperforms state-of-the-art baseline methods.
\label{tab:cholecseg}
    }
    \resizebox{\textwidth}{!}{ 
        \begin{tabular}{@{}l|ccccccc@{}}
        \toprule
            \multirow{2}{*}{Method}& \multicolumn{1}{c}{\parbox{1.5cm}{\centering 01\_00080}} & \multicolumn{2}{c}{01\_00240} & \multicolumn{2}{c}{01\_15019} & \multicolumn{2}{c}{12\_15750} \\
            & Liver  & Liver & Grasper& \parbox{1.8cm}{\centering Abdominal \\ Wall}& Grasper & Fat & \parbox{1.8cm}{\centering L-hook \\Electrocautery}\\
            \midrule
            CLIP~\cite{radford2021learning} & 64.96 & 57.52 &  4.69& 15.94 & 4.1 & \colorbox{yellow!50}{26.59} & 5.34 \\
            SurgVLP~\cite{yuan2025learning} & 65.07 & 57.04 & 6.01 & \colorbox{yellow!50}{32.99} & 5.91 & 8.03 & 6.14 \\
            CAT-Seg~\cite{cho2024cat} & \colorbox{yellow!40}{89.82} & \colorbox{yellow!40}{79.06} & \colorbox{orange!40}{65.78} & \colorbox{orange!40}{97.24} & \colorbox{orange!40}{68.83} & \colorbox{orange!50}{74.84} & \colorbox{orange!40}{73.29} \\
            Medical SAM3 (3D)~\cite{jiang2026medical} & 8.65 & 6.79 & 8.01 & 0.00 & 0.02 & 0.09 & \colorbox{yellow!40}{69.70} \\
            Medical SAM3 (2D)~\cite{jiang2026medical} & 4.99 & 1.66 & 1.06 & 0.00 & 8.61 & 1.69 & 39.92 \\
            SAM3~\cite{carion2025sam} & \colorbox{orange!50}{92.57} & \colorbox{orange!50}{92.50} & \colorbox{yellow!40}{30.26} & 27.09 & \colorbox{yellow!50}{5.92} & 1.20 & 63.17 \\
            \midrule

            \textbf{Ours} & \colorbox{red!40}{96.92} & \colorbox{red!40}{97.31} & \colorbox{red!40}{90.52} & \colorbox{red!40}{98.13} & \colorbox{red!40}{86.55} & \colorbox{red!40}{83.47} & \colorbox{red!40}{94.53} \\
            \bottomrule
        \end{tabular}
    }


\end{table}
\begin{table}[!t]
\centering%

    \caption{
    \textbf{Quantative results on the EndoVis18~\cite{allan20202018} dataset}.
  We highlighted \colorbox{red!40}{first}, \colorbox{orange!50}{second}, and \colorbox{yellow!50}{third} values. We compare the mIoU scores (\%) for segmentation. The performance of our proposed method surpasses state-of-the-art baselines.
\label{tab:endovis}
    }
\resizebox{\textwidth}{!}{
        \begin{tabular}{@{}l|cccccc@{}}
            \toprule
            \multirow{3}{*}{Method}& \multicolumn{2}{c}{\parbox{1.5cm}{\centering Seq\_5}} & \multicolumn{3}{c}{Seq\_9} \\
            & \parbox{1.8cm}{\centering instrument\\-wrist} & \parbox{1.9cm}{\centering kidney\\-parenchyma} & \parbox{1.8cm}{\centering instrument\\-shaft}& \parbox{1.6cm}{\centering instrument\\-wrist}& \parbox{1.8cm}{\centering instrument\\-clasper} \\
            \midrule
            CLIP~\cite{radford2021learning} & 5.17 & {59.17} & 15.55 & 4.1 & {15.94} \\
            SurgVLP~\cite{yuan2025learning} & {11.2} & 21.26 & 25.3 & 4.73 & 3.53 \\
            CAT-Seg~\cite{cho2024cat} & \colorbox{orange!40}{43.29} & \colorbox{orange!40}{71.98} & {20.79} & \colorbox{orange!40}{24.34} & \colorbox{orange!40}{38.94} \\
            Medical SAM3 (3D)~\cite{jiang2026medical} & 14.02 & 9.10 & 32.93 & 4.46 & \colorbox{yellow!50}{34.58} \\
            Medical SAM3 (2D)~\cite{jiang2026medical} & 20.45 & 13.32 & \colorbox{orange!50}{48.29} & 3.92 & 26.31 \\
            SAM3~\cite{carion2025sam} & \colorbox{yellow!50}{23.68} & \colorbox{yellow!40}{61.45} & \colorbox{yellow!50}{34.36} & \colorbox{yellow!50}{6.74} & 4.64 \\
            \midrule

            \textbf{Ours} & \colorbox{red!40}{64.39} & \colorbox{red!40}{73.53} & \colorbox{red!40}{73.90} & \colorbox{red!40}{39.18} & \colorbox{red!40}{40.51} \\
            \bottomrule
        \end{tabular}
    }


\end{table}

\subsection{Results}
We adopt per-class mIoU (\%) to quantify segmentation performance, and compare our LoRA-adapted SAM3 against representative segmentation baselines, including CLIP~\cite{radford2021learning}, SurgVLP~\cite{yuan2025learning}, CAT-Seg~\cite{cho2024cat}, SAM3 and both medical-domain pre-trained 2D/3D versions of Medical SAM3~\cite{jiang2026medical}.
Tab.~\ref{tab:cholecseg} summarizes the results on \textit{CholecSeg8k}.
Zero-shot SAM3 performs well on general anatomical structures but degrades severely on fine-grained surgical classes outside its pre-training scope.
Though trained on extensive medical data, Medical SAM3 underperforms zero-shot SAM3 on surgical tasks, as full fine-tuning across 33 multi-modal datasets dilutes its specialized surgical features.
In comparison, our method preserves pre-trained representations while still adapting to surgical concepts.
Across the multiple sequences it brings substantial gains over zero-shot SAM3 and Medical SAM3 on most surgical categories, with consistent per-class improvements on visually subtle structures such as grasper, fat and L-hook electrocautery.
Quantitative evaluations on \textit{EndoVis18} (Tab.~\ref{tab:endovis}) also show that SAM3 struggles with surgical instrument concepts without adaptation, whereas our LoRA tuning lifts segmentation accuracy across almost all instrument subcategories and both test sequences.
The adapted model obtained an overall mIoU of $26.9\%$ and mDice of $37.0\%$, outperforming both vanilla SAM3 and the Medical SAM3 variants on every test sequence and confirming that lightweight parameter-efficient adaptation transfers across procedures.

\begin{figure*}[t]
  \begin{center}
    \centerline{\includegraphics[width=1\columnwidth]{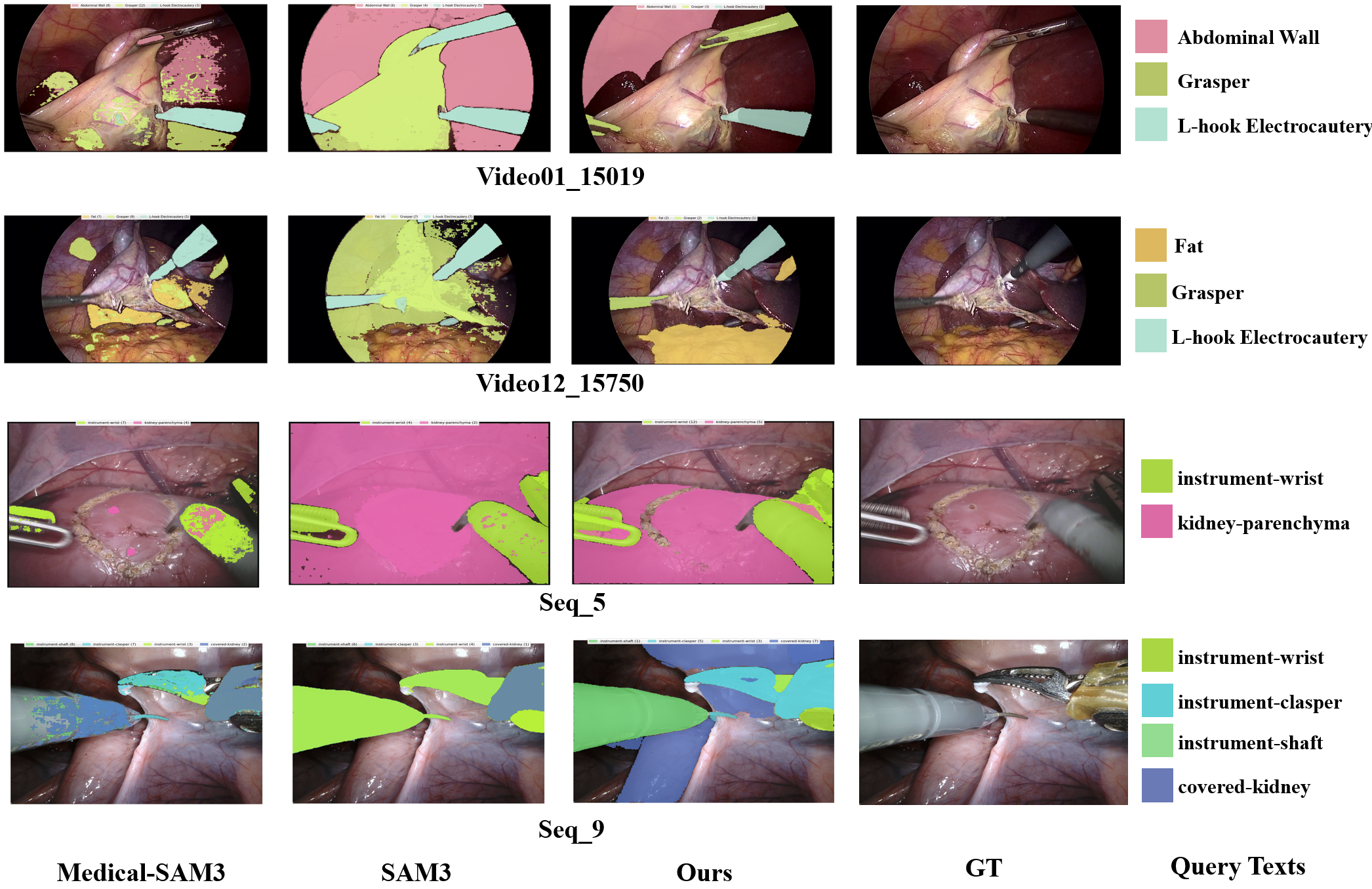}}
    \caption{
      Qualitative comparison of our method and other baselines on representative frames from \textit{CholecSeg8k} and \textit{EndoVis18}.
    }
    \label{fig:qualitative_result}
  \end{center}
\end{figure*}

Fig.~\ref{fig:qualitative_result} provides qualitative comparisons between SAM3, Medical SAM3, and our method on typical surgical frames.
Among all evaluated methods, Medical SAM3, which produces heavily distorted tissue contours, generates spurious structures and misses large liver and connective tissue regions, even underperforming vanilla SAM3.
This may suggest that full-parameter fine-tuning on medical data may damage the foundation model's pre-trained representations.
In contrast, our LoRA-adapted model preserves accurate tissue contours while precisely capturing fine-grained instrument boundaries.

\begin{figure*}[t]
  \centering
  \includegraphics[width=0.9\textwidth]{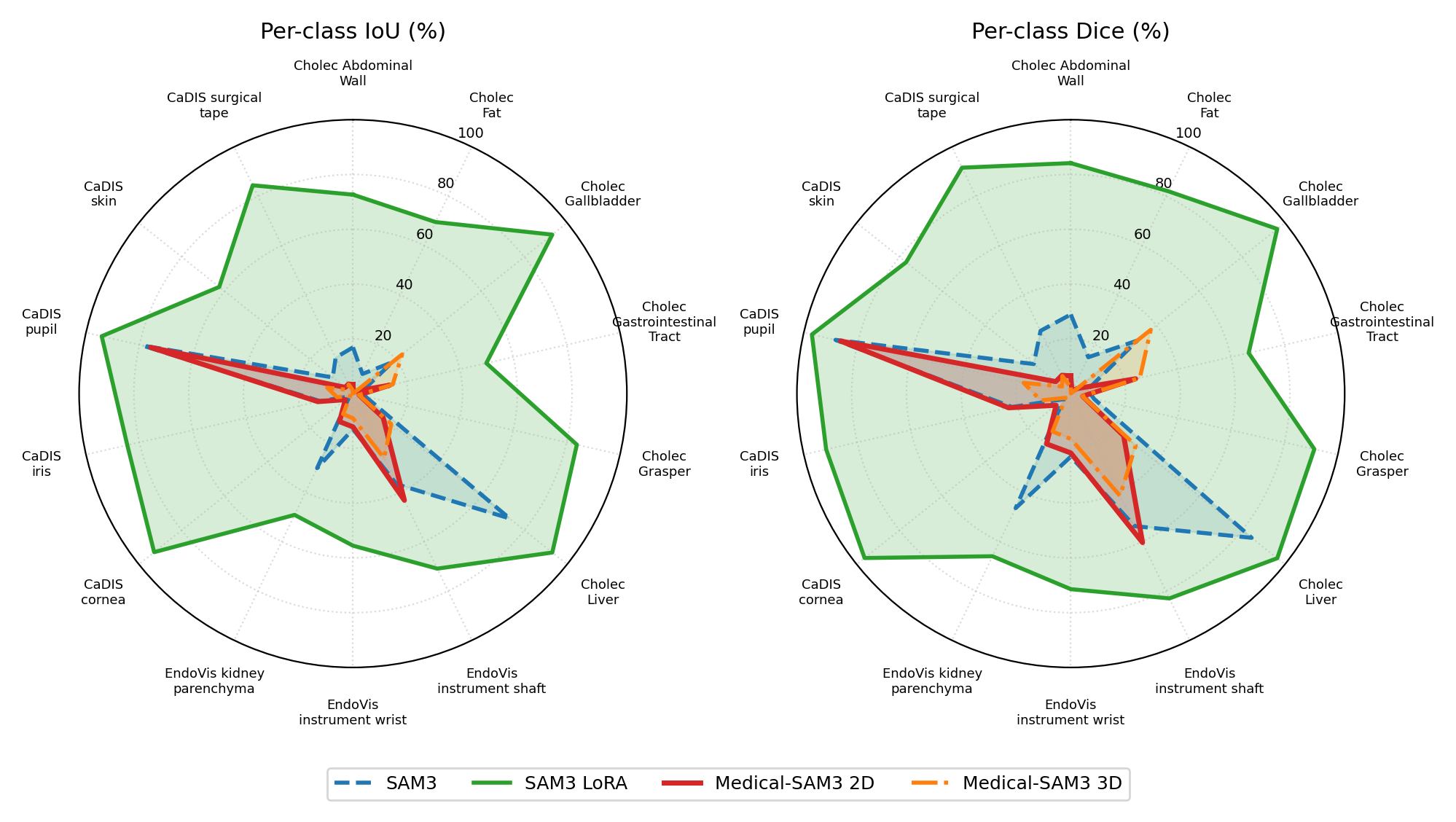}
  \caption{
    Per-class segmentation results on \textit{CholecSeg8k}, \textit{EndoVis18} and \textit{CaDISv2} test splits. Left: mean IoU (\%); Right: mean Dice (\%).
    Axes cover a curated subset of tissue and instrument categories to streamline visual interpretation.
  }
  \label{fig:radar}
\end{figure*}

We further summarise the comparison in Fig.~\ref{fig:radar}.
Our lightweight adaptation of SAM3 achieves consistent performance gains over both baselines.
Zero-shot SAM~3 (blue dashed) performs poorly on most classes, with scores mostly below 40\%.
Between the two fully medical-domain pre-trained baselines, Medical-SAM3 2D (red) delivers moderate performance and works better on \textit{CaDISv2} dataset, while Medical-SAM3 3D (orange) suffers dramatic accuracy declines below 30\% for the majority of anatomical categories.

\subsection{Application}
\label{sec:discussion}

\begin{figure*}[t]
  \centering
  \includegraphics[width=1\textwidth]{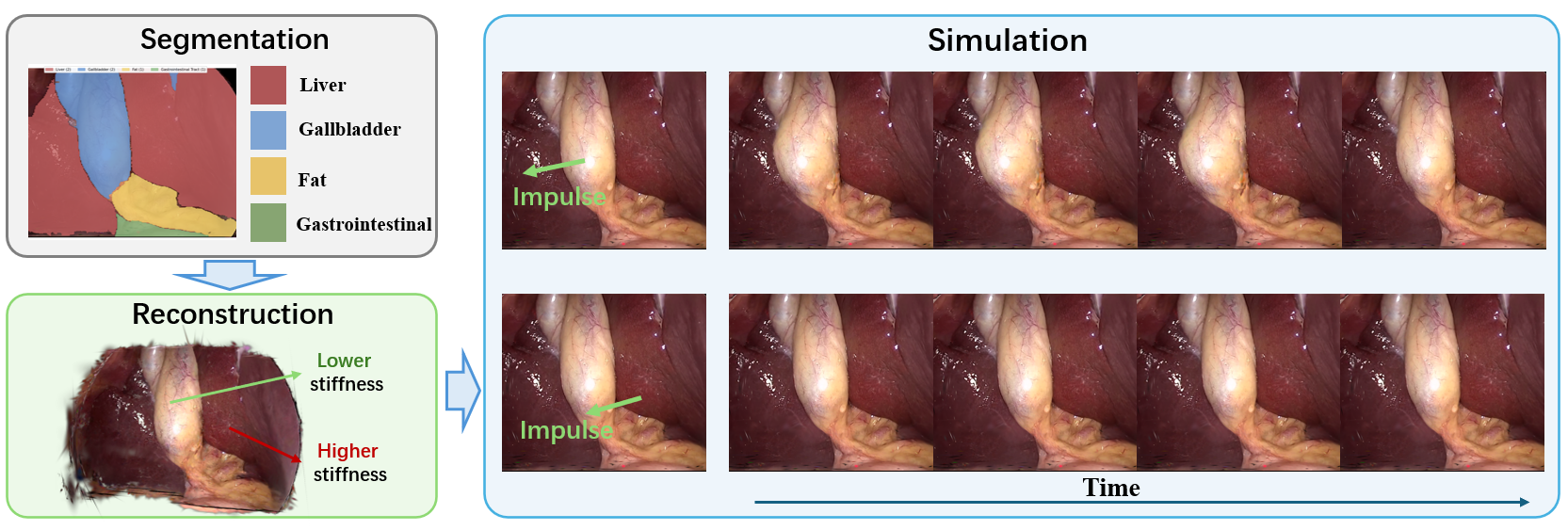}
  \caption{
    From segmentation to downstream applications. \textbf{Left:} masks predicted by our method supply region-level tissue labels to assign unique physical material parameters in surgical simulation.  \textbf{Right:} Applying the same force on tissues with different stiffness (e.g.\ soft \emph{gallbladder} vs.\ stiff \emph{liver}) produces visibly different deformation.
  }
  \label{fig:simulation}
\end{figure*}


Segmentation is a prerequisite for several downstream robotic-surgery tasks, including scene reconstruction and physics-based simulation.
As illustrated in Fig. \ref{fig:simulation}, we adopt a Gaussian Splatting pipeline~\cite{huang2024endo} to reconstruct surgical scenes, and leverage a surgical physics simulation framework~\cite{xie2024physgaussian,liu2026endogsim} to validate the practical applicability of our method.

Our pipeline bridges reconstruction and physics simulation through three key steps.
First, we reconstruct the surgical scene using Gaussian Splatting~\cite{kerbl20233d,huang2024endo} to obtain a representation of the surgical environment.
Second, our method generates class-aware segmentation masks from text prompts on each frame, where each anatomical class (e.g., gallbladder, liver) is identified by a distinct mask region.
These per-frame instance label maps drive region-wise material assignment: each connected component corresponds to one anatomical structure with class-specific physical parameters (e.g., Young's modulus and Poisson's ratio).
Third, to achieve physics-aware dynamic simulation, we initialize each anatomical region with material properties derived from medical literature, then simulate via differentiable Material Point Method (MPM)~\cite{hu2018moving}.
Under the same external forces in simulations, the soft gallbladder undergoes larger deformation than the relatively stiff liver, as illustrated on the right side of Fig. \ref{fig:simulation}.
This demonstrates that concept segmentation masks yield realistic and physically meaningful deformation cues for downstream robotic surgery tasks.


\section{Conclusion}

We introduce a parameter-efficient LoRA adaptation of SAM3 for surgical scene segmentation.
To unify prompts across diverse surgical data, we build a unified input representation that aligns class labels into one consistent space, supporting training with a universal LoRA weight.
By attaching low-rank adapters to the prompt encoder, detector, and tracker while keeping the foundation backbone frozen, the proposed method updates only \(0.98\%\) of SAM3's parameters.
It limits peak training GPU memory to merely 9 GB, delivering over an 80\% reduction in memory overhead compared with full-parameter fine-tuning, making our lightweight scheme highly suitable for deployment in resource-limited clinical and surgical environments.
Experiments on \textit{CholecSeg8k}, \textit{EndoVis18}, and \textit{CaDISv2} show that our method achieves substantial mIoU gains over zero-shot SAM3 and over the fully fine-tuned Medical SAM3 baselines. 

Although our framework achieves strong segmentation results and reliable downstream simulation outputs, there are still multiple dimensions for optimization.
We will further explore memory-light training schemes to lower GPU memory consumption and maximize the efficiency of the base model. Additionally, we also plan to expand our image-based fine-tuning framework to support surgical video sequences.
We will also strengthen end-to-end links with downstream pipelines for real-world clinical robotic tasks, allowing our parameter-efficient adaptation of SAM3 to produce reliable segmentation results for surgical reconstruction and physics-based simulation for surgical robotics policies.



\bibliographystyle{splncs04}
\bibliography{Paper-25}





\end{document}